\documentclass[fleqn,10pt]{olplainarticle}
\usepackage{algorithm}
\usepackage{algpseudocode}
\usepackage[utf8]{inputenc}

\usepackage{tikz}
\usepackage{url}
\usetikzlibrary{shapes.geometric, arrows, positioning}
\usepackage[obeyFinal]{easy-todo}
\usepackage[skip=0.5ex]{subcaption}

\title{Assessing Dataset Quality Through Decision Tree Characteristics in Autoencoder-Processed Spaces}

\author[1]{Szymon Mazurek}
\author[2]{Maciej Wielgosz}
\affil[1]{ACK Cyfronet AGH, ul. Nawojki 11, 30-950 Krakow, POLAND}
\affil[2]{AGH University of Science and Technology, 30 Mickiewicza Ave., 30-059 Krakow, POLAND}


\begin{abstract}
In this paper, we delve into the critical aspect of dataset quality assessment in machine learning classification tasks. Leveraging a variety of nine distinct datasets, each crafted for classification tasks with varying complexity levels, we illustrate the profound impact of dataset quality on model training and performance. We further introduce two additional datasets designed to represent specific data conditions - one maximizing entropy and the other demonstrating high redundancy. Our findings underscore the importance of appropriate feature selection, adequate data volume, and data quality in achieving high-performing machine learning models. To aid researchers and practitioners, we propose a comprehensive framework for dataset quality assessment, which can help evaluate if the dataset at hand is sufficient and of the required quality for specific tasks. This research offers valuable insights into data assessment practices, contributing to the development of more accurate and robust machine learning models.
\end{abstract}

\begin{document}

\flushbottom
\maketitle
\thispagestyle{empty}

\section*{Introduction}

Data is the lifeblood of machine learning. The quality, size, and diversity of the dataset largely determine the effectiveness of a machine learning algorithm, particularly in classification tasks. Being able to properly assess the quality of a dataset is therefore fundamental to the success of any machine learning project.

In the world of machine learning, data is both the input and the teacher. When training a model, it is the dataset that instructs the model on the relationships it needs to learn. Given this, the importance of the dataset cannot be overstated. The adage "garbage in, garbage out" is particularly apt in this context. If the data used for training is of poor quality, biased, or unrepresentative of the problem domain, the resulting model will inevitably be flawed, leading to poor performance and potentially harmful decisions.

The quantity of data is equally crucial. Generally, larger datasets allow models to learn more nuanced relationships, and make them more robust against overfitting. However, gathering massive amounts of data can be costly and time-consuming. Therefore, knowing when we have enough data is a valuable skill. Gathering more data when it isn't needed is wasteful, whereas not gathering enough could mean missing out on important insights or creating an underperforming model.

In addition, the type of classification task has significant implications for the quality and quantity of data needed. For example, a binary classification task may require less data than a multi-class problem, and imbalanced classes can significantly affect model performance if not appropriately addressed. Therefore, it is crucial to understand the specific data requirements for the task at hand.

One significant issue in machine learning classification tasks occurs when the dataset does not contain the right or sufficient features to enable high-quality model training. Feature selection is as crucial as the quantity and quality of data, as it directly affects the model's ability to learn and make accurate predictions. When the dataset lacks appropriate features, the model may suffer from underfitting, where it fails to capture the underlying patterns in the data. In such a case, even the most advanced algorithms would struggle to achieve high accuracy, as indicated by metrics like F1 score. Furthermore, the absence of relevant features may force the model to overly rely on available but less significant features, leading to overfitting and poor generalization to unseen data. Therefore, an effective dataset quality assessment framework should consider not just the quantity and quality of data, but also the appropriateness and informativeness of the features contained within the data. This comprehensive approach to data assessment can help in creating models with higher predictive accuracy and robustness.

In summary, the ability to evaluate dataset quality is of paramount importance to machine learning, especially in classification tasks. A well-designed dataset quality assessment framework can help researchers and practitioners understand whether they have enough data and whether the data is of sufficient quality for their specific tasks. In this paper, we propose such a framework, discuss its components and illustrate its application in several real-world scenarios.

\section*{Methods and Materials}

\subsection{Datasets used}

For the experimental study, we employed a diverse assortment of nine datasets, each meticulously curated to serve classification tasks with varying degrees of complexity. These datasets encapsulated images distributed among multiple classes. Two specific datasets, the 'expert' dataset and the 'pitbull' dataset, as detailed in our previous publication \cite{DogAgeMazurek}, were chosen as representative instances of deficient datasets. These datasets were characterized by their inability to deliver high performance for deep learning classifiers designed to differentiate between distinct dog age groups - a stark reflection of their inherent limitations in fostering effective model training.

The following five datasets - MNIST, FMNIST, CIFAR10, Tiny ImageNet, and Stanford Dogs - have gained recognition as benchmark datasets for computer vision tasks reliant on deep learning. These datasets were selected for their proven ability to enhance the sophistication of trained classifiers, enabling them to achieve, if not exceed, human-level performance. These datasets, diverse in their representation of complex objects, were considered archetypal of 'good' datasets.

To complement the above sets, we developed two additional datasets with unique characteristics. The first was generated through a process of random selection, pulling pixel values and class assignments from a Gaussian distribution. This dataset was designed to maximize entropy, thereby representing a dataset devoid of discernible patterns or information useful for classifier training. The second dataset was built from augmenting ten arbitrarily selected images from the CIFAR10 dataset, each representing a unique class. The augmentation process incorporated random rotation and arbitrary modifications to attributes such as saturation, sharpness, hue, and contrast. The intent was to model a dataset demonstrating high redundancy - while the samples varied, they largely conveyed repetitive information due to their origin from the same source images.

Each dataset used in this study underwent a standardization process, where the samples were resized to a unified resolution of 128x128 via bilinear interpolation. An extensive description of these datasets, including their unique characteristics and classifications, can be found in Table \ref{tab:datasets} and in Table \ref{tab:datasets_details} in the Appendix section.

\begin{table}[h]
\centering
\begin{tabular}{|c|c|c|}
\hline
\textbf{Dataset} & \textbf{Description} & \textbf{Category} \\
\hline
Expert & Poor performance in age group classification & Bad \\
\hline
Pitbull & Poor performance in age group classification & Bad \\
\hline
MNIST & Benchmark for digit recognition tasks & Good \\
\hline
FMNIST & Benchmark for fashion item recognition tasks & Good \\
\hline
CIFAR10 & Benchmark for object recognition tasks & Good \\
\hline
Tiny ImageNet & Benchmark for image recognition tasks & Good \\
\hline
Stanford Dogs & Benchmark for dog breed recognition tasks & Good \\
\hline
Random Gaussian & High entropy, no discernible patterns & Special \\
\hline
Augmented CIFAR10 & High redundancy, repetitive information & Special \\
\hline
\end{tabular}
\caption{Overview of datasets used in the study}
\label{tab:datasets}
\end{table}

\subsection{Experimental setup}

The experimental procedure was undertaken within a Python 3.10 environment, utilising PyTorch 1.13 and PyTorch Lightning 1.9.0 for autoencoders implementation and Scikit-learn 1.2.1 for a binary tree classifier. The preliminary step involved training a model for dataset-specific data compression. Autoencoders were employed for this purpose, selected for their relative simplicity in terms of hyperparameter specification compared to other models such as Variational Autoencoders (VAE). This strategy aimed at reducing the experiment-controlled parameters complexity, allowing a sharper focus on inherent dataset properties.

The autoencoders were trained using the Adam optimizer with a learning rate set at 0.0001 to minimize the Mean Squared Error (MSE) between input and output. The batch size was predetermined at 256. The data was divided into training and validation subsets in a dataset-dependent manner. The MNIST, FMNIST, and CIFAR10 datasets used validation sets provided by the original authors. In contrast, a random partition was made for the remaining datasets, setting aside 20\% as a validation subset. 

This subset was employed to drive an early stopping mechanism halting training if no improvement in Mean Absolute Error (MAE) on validation data was observed within the last three epochs. Six autoencoders were trained for each dataset, with their characteristics outlined in Table \ref{tab:datasets}. 

Post training, the dataset was embedded using the encoder segment and saved. Random samples were subsequently selected from the raw dataset or from one of the previously derived embedding sets. This sampling continued until the pre-determined number of samples in each class was achieved. 

Each drawn subset was used for binary tree classifier training. The Gini impurity measure was used to evaluate the quality of splits during the tree training. There were no restrictions on the tree's depth or leaf count. After training, we recorded the maximum depth and leaf count of the tree.

To explore sample size effects on the obtained metrics, subsets were crafted for tree training by sampling 10, 25, 50, 75, or 100 samples per class . 

All experiments were conducted on HPC cluster. For training of autoencoders,  4 AMD EPYC 7742 64-Core CPUs and 4 Nvidia A-100 GPU with CUDA 11.7 drivers were used. The embedding was performed using the same number of CPUs and single GPU. For training of binary tree classifier, only 2 CPUs were used.

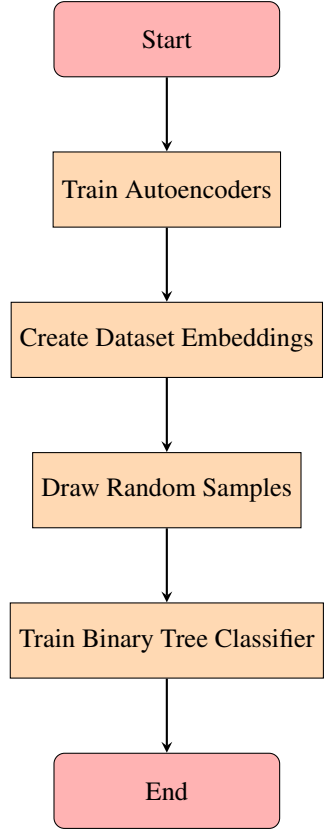
\begin{figure}[h]
\centering
\begin{tikzpicture}[
    node distance=2cm,
    startstop/.style={rectangle, rounded corners, minimum width=3cm, minimum height=1cm, text centered, draw=black, fill=red!30},
    process/.style={rectangle, minimum width=3cm, minimum height=1cm, text centered, draw=black, fill=orange!30},
    arrow/.style={thick,->,>=stealth}
]

\node (start) [startstop] {Start};
\node (proc1) [process, below of=start] {Train Autoencoders};
\node (proc2) [process, below of=proc1] {Create Dataset Embeddings};
\node (proc3) [process, below of=proc2] {Draw Random Samples};
\node (proc4) [process, below of=proc3] {Train Binary Tree Classifier};
\node (stop) [startstop, below of=proc4] {End};

\draw [arrow] (start) -- (proc1);
\draw [arrow] (proc1) -- (proc2);
\draw [arrow] (proc2) -- (proc3);
\draw [arrow] (proc3) -- (proc4);
\draw [arrow] (proc4) -- (stop);

\end{tikzpicture}
\caption{Flowchart of the experimental procedure. The process starts with training autoencoders, followed by creating dataset embeddings. Then, random samples are drawn from the datasets, and a binary tree classifier is trained on these samples. The process ends once the binary tree classifier is trained.}
\label{fig:exp_setup}
\end{figure}

To formalize the presented approach, let's define the following notations:

\begin{align*}
D & : \text{Dataset} \\
D_{tr} & : \text{Training set} \\
D_{val} & : \text{Validation set} \\
C & : \text{Number of classes} \\
N & : \text{Total instances in the dataset} \\
N_{c} & : \text{Instances in class } c, \text{ where } c = 1,2,...,C \\
p(c) & : \text{Probability of class } c, p(c) = \frac{N_{c}}{N} \\
AE & : \text{Autoencoder} \\
E & : \text{Encoder part of the autoencoder} \\
f_{AE} & : \text{Reconstructed output of the autoencoder} \\
MSE_{AE} & : \text{Mean Squared Error between input images and } f_{AE} \\
BT & : \text{Binary tree classifier} \\
GI & : \text{Gini impurity} \\
n_{leaf} & : \text{Leaf count in the tree} \\
d_{max} & : \text{Maximum depth of the tree}
\end{align*}

The entropy of a dataset, measuring its information content, is defined as:

\begin{equation}
H(D) = - \sum_{c=1}^{C} p(c) \log_{2} p(c)
\end{equation}

The following equations outline the main steps of our experimental setup:

1. The autoencoder is trained by minimizing the Mean Squared Error:

\begin{equation}
\min_{AE} MSE_{AE}(D_{tr}, f_{AE}(D_{tr}))
\end{equation}

2. The dataset is then embedded using the encoder part of the autoencoder:

\begin{equation}
D_{emb} = E(D)
\end{equation}

3. The binary tree classifier is trained by minimizing the Gini impurity:

\begin{equation}
\min_{BT} GI(D_{emb}, BT(D_{emb}))
\end{equation}

Additionally, let's assume we have a trained binary tree classifier. We can express the number of leaves and the maximum depth of the tree as $n_{leaf}(BT)$ and $d_{max}(BT)$, respectively.

\section*{Results}
Metrics extracted for the trees trained on data from every datasets were grouped together for between-datasets comparison. The presented results were obtained by summing the metrics for trees trained on subsets with different number of examples per class. The results are visible in Figure \ref{fig:dataset_bar}. In \ref{fig:dataset_trend}, the resulting metric values for particular sample sizes are shown.

\begin{figure}
    \centering

    \centering
    \begin{subfigure}{420pt}
        \includegraphics[scale=1.0]{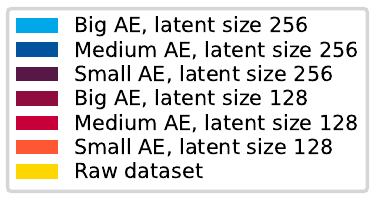}
    \end{subfigure}

    \begin{subfigure}{1\textwidth}
        \includegraphics[scale=0.19]{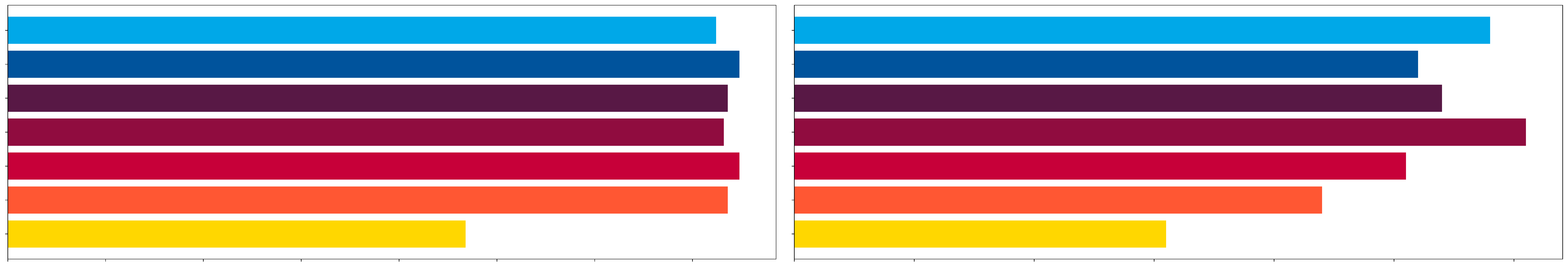}
        \subcaption{Expert dog age}
        \label{fig:dataset_bar1}
    \end{subfigure}

    \begin{subfigure}{1\textwidth}
        \includegraphics[scale=0.19]{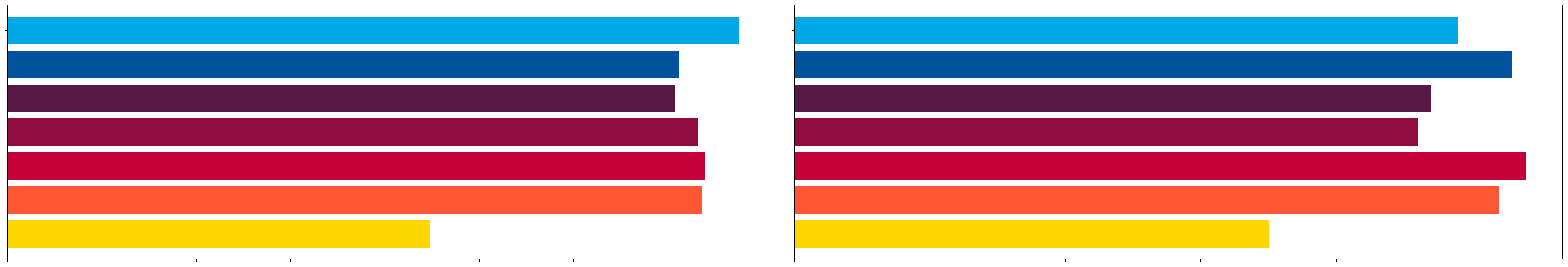}
        \subcaption{Pitbull dog age}
        \label{fig:dataset_bar2}
    \end{subfigure}

    \begin{subfigure}{1\textwidth}
        \includegraphics[scale=0.19]{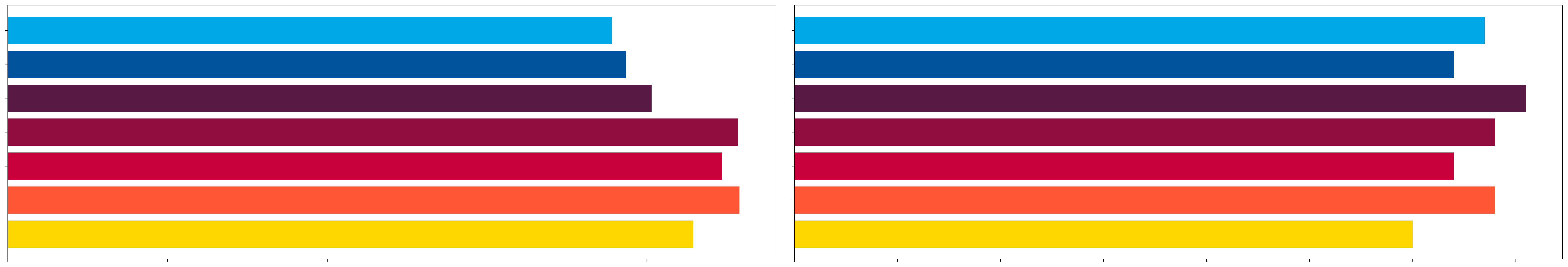}
        \subcaption{MNIST}
        \label{fig:dataset_bar3}
    \end{subfigure}
    
    \begin{subfigure}{1\textwidth}
        \includegraphics[scale=0.19]{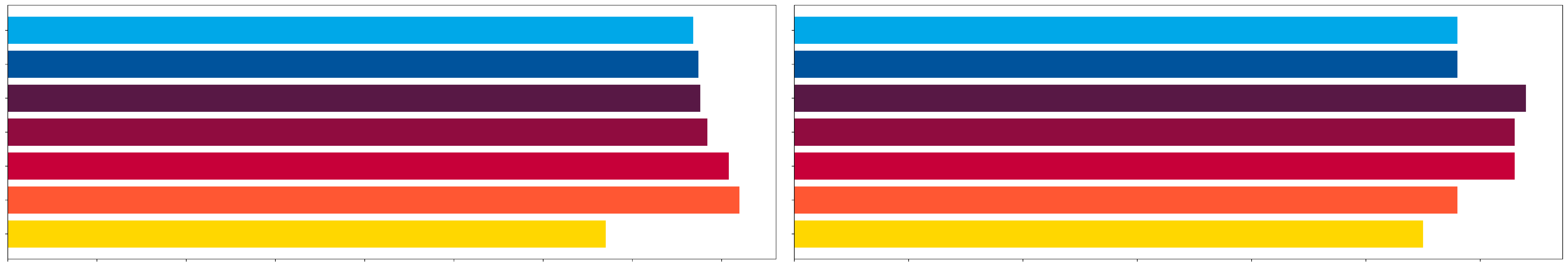}
        \subcaption{FMNIST}
        \label{fig:dataset_bar4}
    \end{subfigure}

    \begin{subfigure}{1\textwidth}
        \includegraphics[scale=0.19]{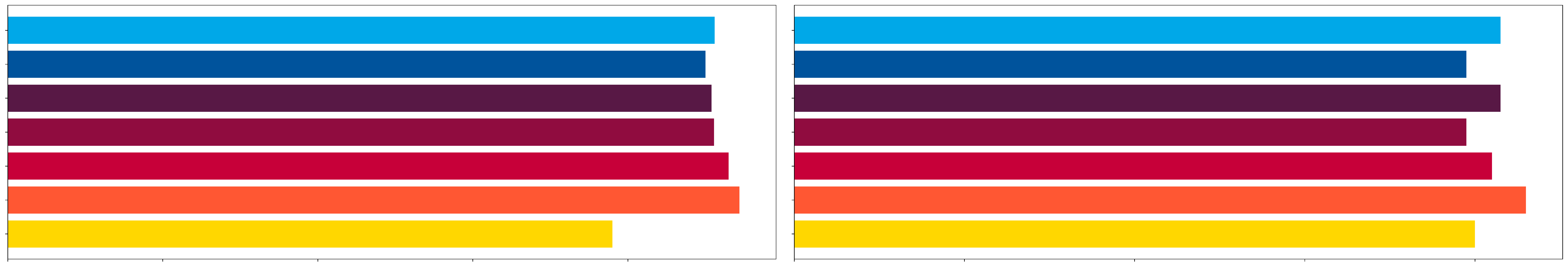}
        \subcaption{CIFAR10}
        \label{fig:dataset_bar5}
    \end{subfigure}

    \begin{subfigure}{1\textwidth}
        \includegraphics[scale=0.19]{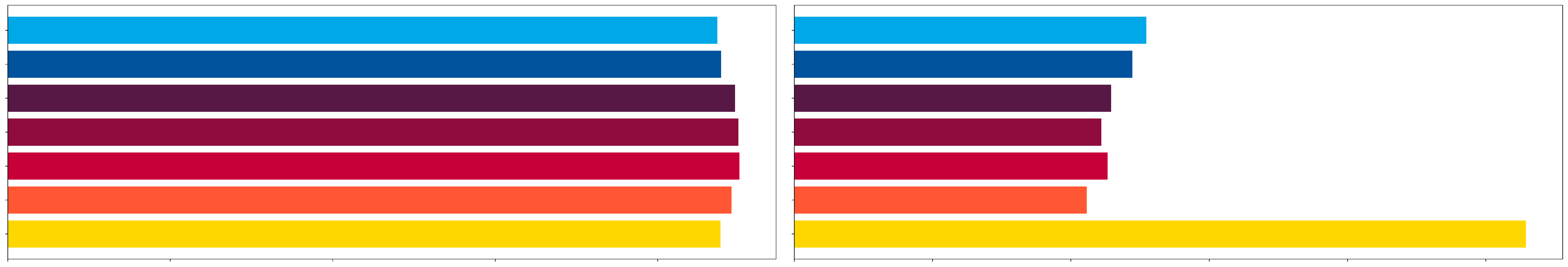}
        \subcaption{Stanford Dogs}
        \label{fig:dataset_bar6}
    \end{subfigure}

\caption[]{Metric values for different datasets, presented as sum of obtained values for every experiment with different per class sample size for each AE embedding scenario}
\end{figure}

\begin{figure}\ContinuedFloat
    \centering
    \begin{subfigure}{420pt}
        \includegraphics[scale=1.0]{legends/legend_bar.pdf}
    \end{subfigure}

    \begin{subfigure}{1\textwidth}
        \includegraphics[scale=0.19]{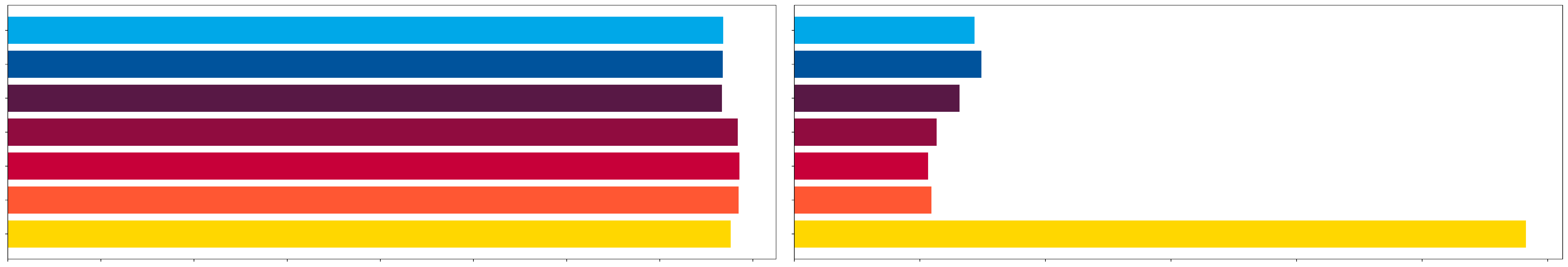}
        \subcaption{Tiny Imagenet}
        \label{fig:dataset_bar7}
    \end{subfigure}
    
    \begin{subfigure}{1\textwidth}
        \includegraphics[scale=0.19]{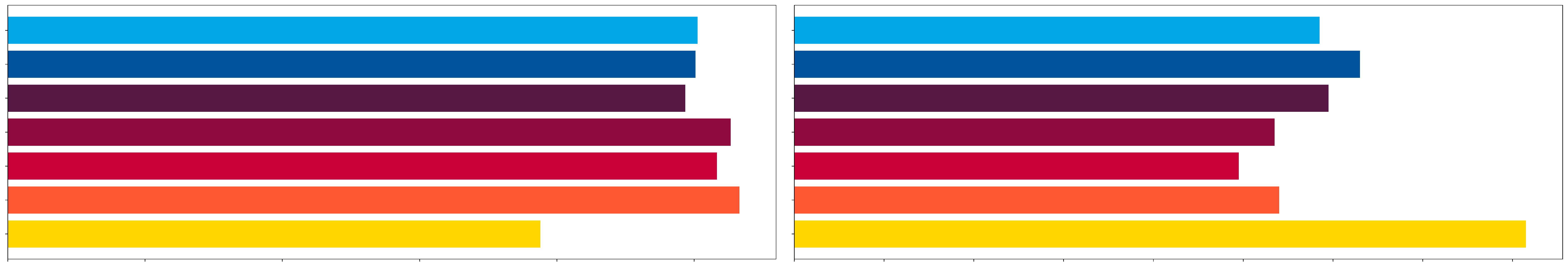}
        \subcaption{Random dataset}
        \label{fig:dataset_bar8}
    \end{subfigure}
    
    \begin{subfigure}{1\textwidth}
        \includegraphics[scale=0.19]{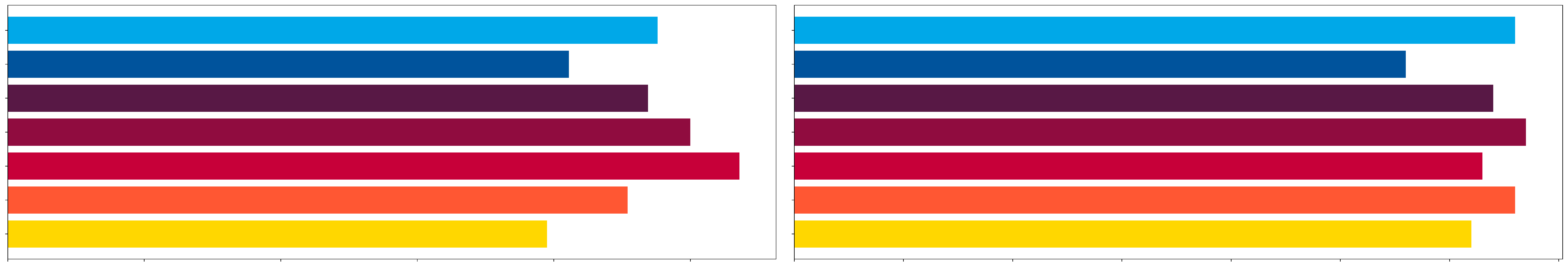}
        \subcaption{Repeated dataset}
        \label{fig:dataset_bar9}
    \end{subfigure}
    \caption[]{Metric values for different datasets, presented as sum of obtained values for every experiment with different per class sample size for each AE embedding scenario (cont.) Every bar corresponds to different AE embedding, as described in the legends. Bar plots in the left column show the number of leaves. Bar plots in the right column show maximum tree depth.}
    \label{fig:dataset_bar}
\end{figure}

\begin{figure}[!h]
    
    \centering
    \begin{subfigure}{420pt}
        \includegraphics[scale=1.0]{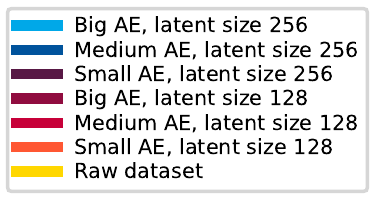}
    \end{subfigure}
    
    \begin{subfigure}{1\textwidth}
        \includegraphics[scale=0.19]{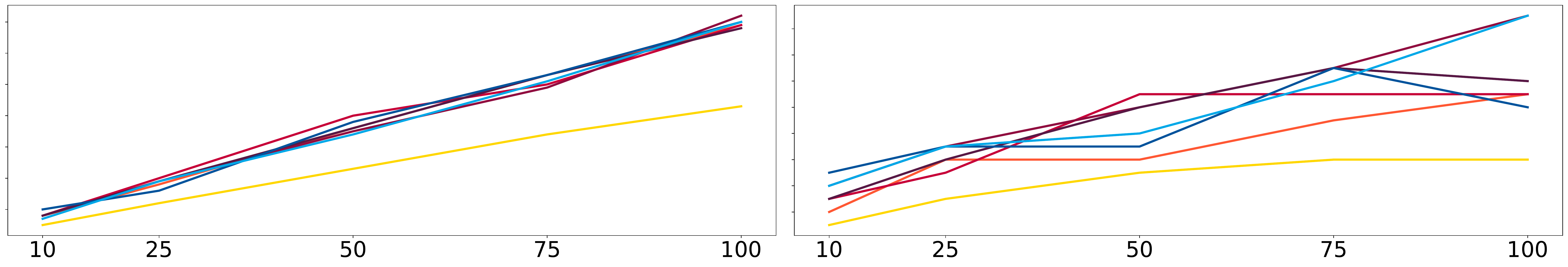}
        \subcaption{Expert dog age}
        \label{fig:dataset_trend1}
    \end{subfigure}

    \begin{subfigure}{1\textwidth}
        \includegraphics[scale=0.19]{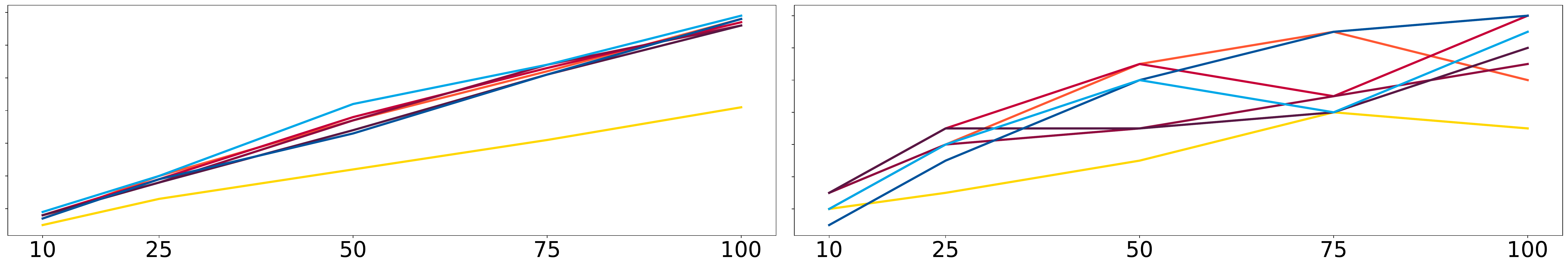}
        \subcaption{Pitbull dog age}
        \label{fig:dataset_trend2}
    \end{subfigure}

    \begin{subfigure}{1\textwidth}
        \includegraphics[scale=0.19]{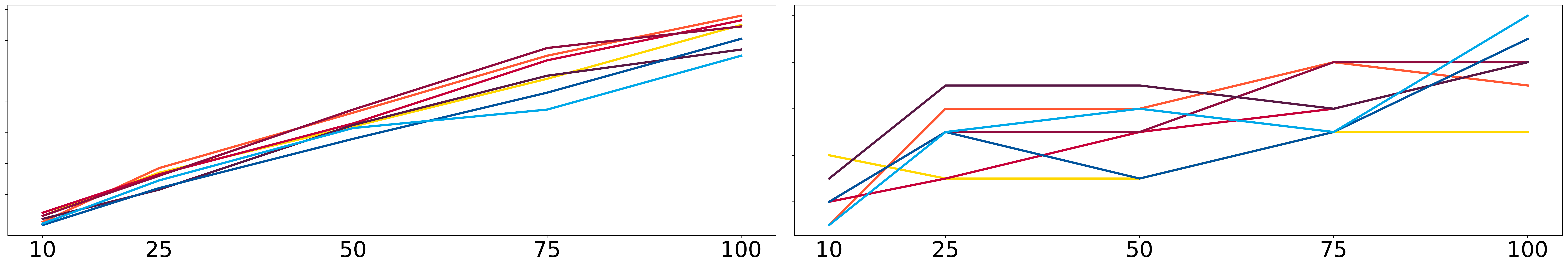}
        \subcaption{MNIST}
        \label{fig:dataset_trend3}
    \end{subfigure}
    
    \begin{subfigure}{1\textwidth}
        \includegraphics[scale=0.19]{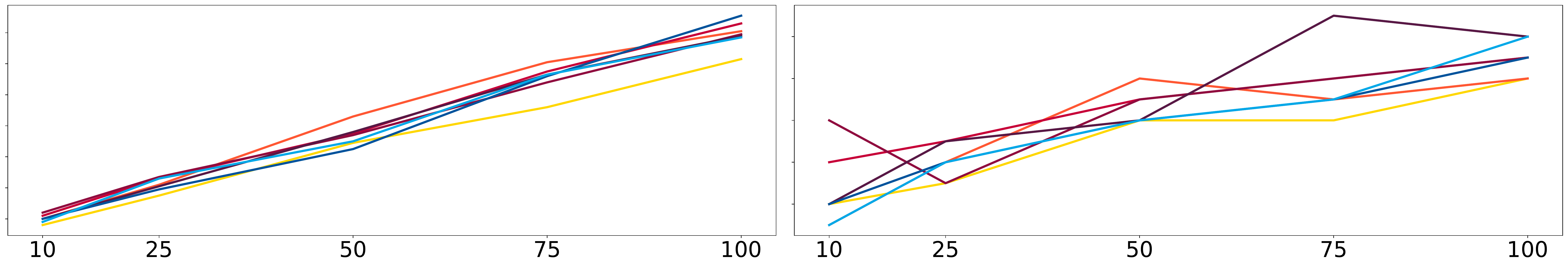}
        \subcaption{FMNIST}
        \label{fig:dataset_trend4}
    \end{subfigure}

    \begin{subfigure}{1\textwidth}
        \includegraphics[scale=0.19]{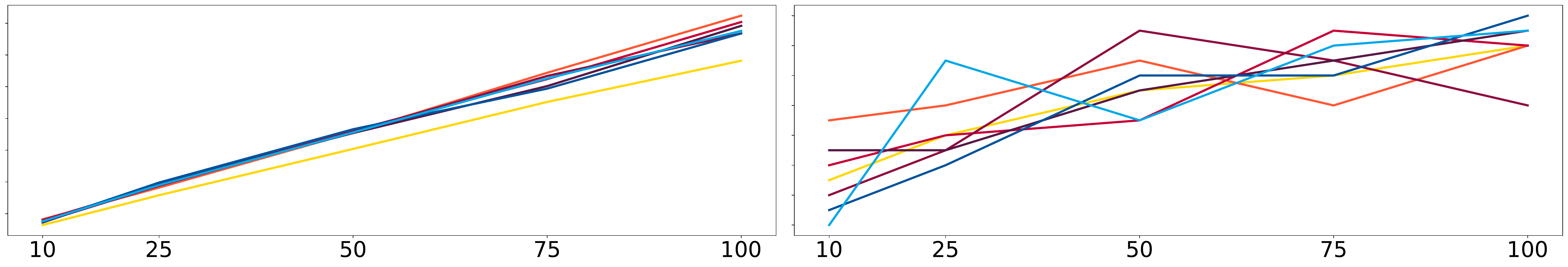}
        \subcaption{CIFAR10}
        \label{fig:dataset_trend5}
    \end{subfigure}

    \begin{subfigure}{1\textwidth}
        \includegraphics[scale=0.19]{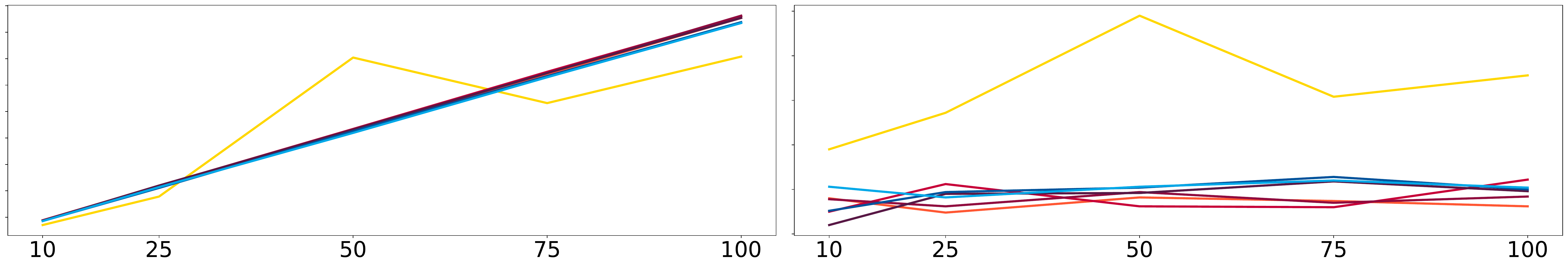}
        \subcaption{Stanford Dogs}
        \label{fig:dataset_trend6}
    \end{subfigure}

\caption[]{Metric values for different datasets for different number of samples per class used.}
\end{figure}

\begin{figure}[!ht]\ContinuedFloat
    \centering

    \centering
    \begin{subfigure}{420pt}
        \includegraphics[scale=1.0]{legends/legend_line.pdf}
    \end{subfigure}

    \begin{subfigure}{1\textwidth}
        \includegraphics[scale=0.19]{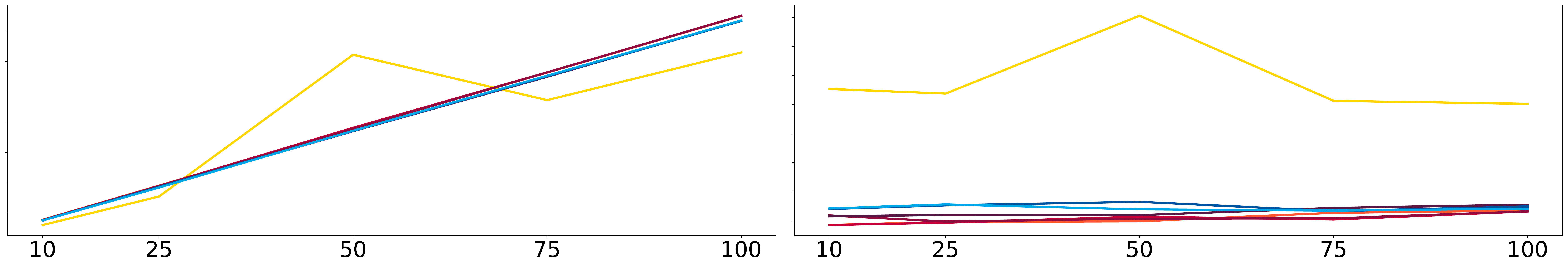}
        \subcaption{Tiny Imagenet}
        \label{fig:dataset_trend7}
    \end{subfigure}
    
    \begin{subfigure}{1\textwidth}
        \includegraphics[scale=0.19]{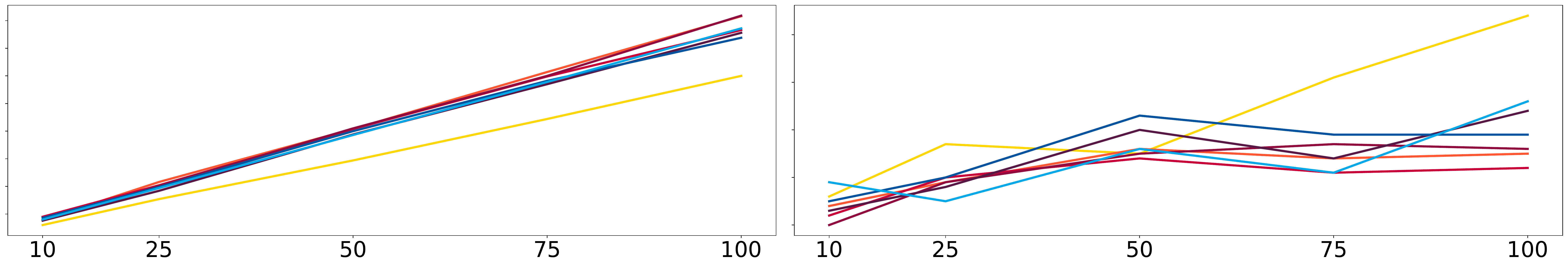}
        \subcaption{Random dataset}
        \label{fig:dataset_trend8}
    \end{subfigure}
    
    \begin{subfigure}{1\textwidth}
        \includegraphics[scale=0.19]{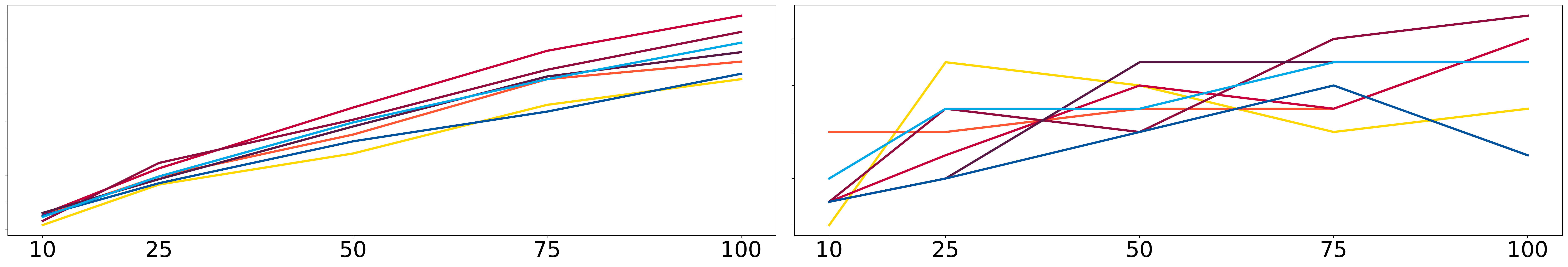}
        \subcaption{Repeated dataset}
        \label{fig:dataset_trend9}
    \end{subfigure}

    \caption[]{Metric values for different datasets for different number of samples per class used (cont.). Every line corresponds to different AE embedding, as described in the legends. The trend seems to be preserved across all sample sizes for "bad" classification datasets.}
    \label{fig:dataset_trend}
\end{figure}

\section*{Discussion}

Our analyses have unveiled noteworthy patterns associated with the employment of decision tree classifiers on diverse datasets. Particularly, for high-quality datasets, a degree of uniformity is discerned when comparing the depth of the decision tree and the number of leaves derived from classifications of the raw and autoencoder-processed datasets. This homogeneity indicates that the influence of the autoencoders on the classification process does not significantly differ from that of the raw data. In simpler terms, both the raw and the autoencoder-processed datasets exhibit similar complexity levels when subjected to decision tree classification, as evidenced by the akin tree depth and leaf count.

Contrastingly, this pattern markedly diverges in the scenario of low-quality datasets. Here, a pronounced discrepancy in the tree depth and the number of leaves is observed when contrasting the classification outcomes from raw datasets and those processed by autoencoders. This difference intimates that the decision tree classifier reacts distinctively to the low-quality data when processed by the autoencoder. This potentially alludes to the introduction or amplification of dataset inconsistencies during the autoencoder's processing phase, leading to a heightening of complexity in the decision tree's structure for the autoencoder-processed dataset, as opposed to the raw data. This observation offers valuable insights for future optimization of data processing pipelines and delineates a promising area for forthcoming research.

In certain circumstances, when dealing with high-quality datasets, decision trees generated from raw data may be significantly more extensive than those resulting from autoencoder-processed data. This observation could be associated with the inherent high dimensionality or complex structure of the raw datasets. In these cases, a decision tree might tend to overfit to the raw data, resulting in a larger tree with more branches and greater depth. The high dimensionality of the raw data might inadvertently encourage the decision tree to find spurious correlations, therefore increasing the complexity of the tree.

On the contrary, when the dataset is pre-processed through an autoencoder, the data's dimensionality is significantly reduced, potentially leading to a simpler, less complex decision tree. The autoencoder could effectively capture the essential features of the data while discarding irrelevant or redundant aspects, consequently enabling a more streamlined decision-making process for the tree classifier. This affirms the benefit of autoencoders in serving as a dimensionality reduction tool, particularly for high-dimensional data, which allows the decision tree to focus on the most salient features for classification without being side-tracked by the potential noise present in the raw, high-dimensional data.

In the case of lower-quality or 'bad' datasets, the discrepancy between the complexity of decision trees trained on raw data versus autoencoder-processed data might not be as pronounced. One plausible explanation for this is the inherent noise and irrelevant patterns present in such datasets. When decision trees attempt to fit on this noisy raw data, they may not necessarily result in more complex structures because the noise and non-informative features do not lend themselves to creating meaningful splits that increase the tree's depth.

Decision trees operate by iteratively creating splits in the dataset that best separate the classes based on a certain metric, such as Gini impurity or information gain. The splits are created based on the values of the different features in the dataset. When a feature effectively separates the classes, it can lead to an increase in the tree's depth as further splits may be necessary down each branch of the tree to fine-tune the classification.

However, when the data is noisy or contains irrelevant features, these do not lead to effective splits that separate the classes. In other words, using these features to create a split in the dataset would not significantly improve the purity of the classes in the resulting subsets. As a result, the decision tree algorithm may not choose to split on these features or may halt further splits down a certain branch, leading to a shallower tree. This is due to the inherent tendency of decision tree algorithms to avoid overfitting by minimizing complexity where the splits do not significantly improve classification performance. Thus, the presence of noise and irrelevant features in a dataset can limit the depth and complexity of the resulting decision tree.

Moreover, when autoencoders are applied to these 'bad' datasets, they may fail to effectively filter out the noise and extract useful features due to the lack of strong, coherent patterns in the data. Therefore, the compressed representation generated by the autoencoder may still contain a significant degree of noise, which leads the decision tree classifier to generate trees of comparable complexity as those produced from the raw data. The autoencoder's inability to significantly enhance the quality of the 'bad' dataset could be seen as the reason why the decision tree's complexity doesn't show noticeable difference whether trained on raw or autoencoder-processed data.

\section*{Conclusions}

In conclusion, our research reveals a notable connection between the quality of datasets, their representation through autoencoders, and the complexity of binary decision tree classifiers trained on these data representations. We found that for high-quality datasets, decision trees trained on autoencoder-compressed data generally exhibited lower complexity compared to trees trained on raw data. This underlines the value of autoencoders in extracting salient features and reducing noise, which assists decision trees in making efficient, meaningful splits.

However, this relationship was not as pronounced in lower quality datasets, suggesting that the benefits of autoencoder compression diminish when the original data is too noisy or lacks informative features. This observation underscores the importance of high-quality data for effective machine learning applications.

The findings of this study not only emphasize the impact of data quality and representation on decision tree complexity but also provide valuable insights into the role of autoencoders as a preprocessing step. It opens up avenues for further exploration on optimizing decision tree classifiers through advanced preprocessing techniques, particularly in the context of varying dataset quality.

\section*{Data availability}

The code and datasets are available in  \url{https://github.com/szmazurek/ds\_assessment} 
\section*{Acknowledgments}

This research was supported in part by the PLGrid infrastructure grant plglaoisi23 on the Athena computer cluster (https://docs.cyfronet.pl/display/~plgpawlik/Athena).

\bibliography{sample}

\section{Appendix}

\begin{algorithm}
\caption{Dataset Quality Evaluation Framework for Machine Learning Classification Tasks}
\begin{algorithmic}[1]
\Require Diverse set of datasets designed for classification tasks
\Ensure Quantitative assessment of dataset quality

\Procedure{Quality Assessment}{}
\State Initialize an assortment of distinct datasets dedicated to various classification tasks
\State Construct two supplemental datasets: one exhibiting maximized entropy and the other showing high redundancy
\State Standardize all samples across datasets to a uniform resolution via bilinear interpolation
\State Implement autoencoder networks, train them on each dataset, and optimize using a loss function such as Mean Squared Error (MSE), comparing input images to reconstructed output
\State Employ the trained autoencoders to generate high-dimensional embeddings for each dataset
\For{each dataset}
    \For{different sample sizes}
        \State Randomly select samples from either the original dataset or one of the generated embedding sets
        \State Employ a binary tree classifier to train on the selected samples, utilizing Gini impurity as a metric for quality assessment of splits during the training
        \State Evaluate the resulting tree, recording the total number of leaves and the maximum depth as indicators of dataset complexity and quality
    \EndFor
\EndFor
\State Aggregate, analyze, and compare results across datasets and sample sizes to determine dataset quality and appropriateness for machine learning classification tasks
\EndProcedure
\end{algorithmic}
\end{algorithm}

\subsection{Dataset detials}

In the table below, a detailed description of used datasets is provided.

\begin{table}[h]
\centering
\begin{tabular}{|l|l|l|l|l|}
\hline
Dataset                                                     & \begin{tabular}[c]{@{}l@{}}Total \\ samples\end{tabular} & \begin{tabular}[c]{@{}l@{}}Total\\ classes\end{tabular} & \begin{tabular}[c]{@{}l@{}}Samples\\ per class\end{tabular}                        & Details                                                                                                                                                                                        \\ \hline
Expert                                                      & 1088                                                     & 3                                                       & \begin{tabular}[c]{@{}l@{}}Young = 223\\ Adult = 370\\ Senior = 495\end{tabular}   & \begin{tabular}[c]{@{}l@{}}Dataset provided by the authors of DogAge \\ Challenge \cite{DogAgeChallengeZamansky}\end{tabular}                                                                                                 \\ \hline
Pitbull                                                     & 3215                                                     & 3                                                       & \begin{tabular}[c]{@{}l@{}}Young = 1071\\ Adult = 1148\\ Senior = 996\end{tabular} & \begin{tabular}[c]{@{}l@{}}Dataset created by the authors of \\ \cite{DogAgeMazurek}\end{tabular}                                                                                            \\ \hline
MNIST                                                       & 70000                                                    & 10                                                      & 7000                                                                               & For more details refer to \cite{MNISTBestOfIeee}                                                                                                                                                                     \\ \hline
FMNIST                                                      & 70000                                                    & 10                                                      & 7000                                                                               & For more details refer to  \cite{FMNISTxiao}                                                                                                                                                                  \\ \hline
CIFAR10                                                     & 60000                                                    & 10                                                      & 6000                                                                               & \begin{tabular}[c]{@{}l@{}}For more details refer to\\  \cite{CIFAR10Krizhevsky} \end{tabular}
                                                                                                                                                                   \\ \hline
\begin{tabular}[c]{@{}l@{}}Tiny\\ ImageNet\end{tabular}     & 98179*                                                   & 200                                                     & 500*                                                                               & \begin{tabular}[c]{@{}l@{}}For more details refer to \cite{TinyImagenetWu} \\ *see caption below\end{tabular}                                                                                                         \\ \hline
\begin{tabular}[c]{@{}l@{}}Stanford\\ Dogs\end{tabular}     & 20580                                                    & 120                                                     & ** & **For more details refer to \cite{StanfordDogsKhosla}                                                                                                                                                                     \\ \hline
\begin{tabular}[c]{@{}l@{}}Random\\ Gaussian\end{tabular}   & 10000                                                    & 10                                                      & 1000 +- 35                                                                         & \begin{tabular}[c]{@{}l@{}}Pixel intensities sampled randomly\\ from normal distribution. Class labels\\ as integers  from 0 to 9, sampled randomly\\  from uniform distribution.\end{tabular} \\ \hline
\begin{tabular}[c]{@{}l@{}}Augmented\\ CIFAR10\end{tabular} & 10000                                                    & 10                                                      & 1000                                                                               & \begin{tabular}[c]{@{}l@{}}One image taken randomly from every class\\ of CIFAR10 dataset, repeated 1000 times with \\ different augmentations.\end{tabular}                                   \\ \hline
\end{tabular}
\caption{Detailed overview of the datasets used. * - For computational efficiency purposes, we decided not to use testing set provided by authors of the dataset. It was therefore not included in any part of experiment. Also, during the data preparation, some of the images in downloaded files were grayscale, therefore they were removed from the training set (total 1821 images removed). }
\label{tab:datasets_details}
\end{table}

\end{document}